\documentclass[conference]{IEEEtran}
\IEEEoverridecommandlockouts
\usepackage{comment}
\usepackage{algorithm}
\usepackage{algorithmic}
\usepackage{afterpage}
\usepackage{placeins}
\usepackage{amsmath,amssymb}
\usepackage{graphicx}
\usepackage{booktabs}
\usepackage{algorithm}
\usepackage{algpseudocode}
\usepackage{tikz}
\usetikzlibrary{positioning,arrows.meta,shapes.geometric}
\usepackage{xcolor}

\usepackage{tabularx}
\usepackage{array}
\usepackage{subcaption}
\usepackage{siunitx}
\usepackage{url}
\usepackage{xcolor}
\providecommand{\rowcolor}[1]{}

\definecolor{rowgray}{gray}{0.92}
\definecolor{bestblue}{RGB}{0,92,184}

\definecolor{ablue}{RGB}{0,92,184}
\definecolor{agreen}{RGB}{0,120,120}
\definecolor{alight}{gray}{0.93}

\newcommand{\bestRMSE}[1]{\textcolor{ablue}{\textbf{#1}}}
\newcommand{\bestAUROC}[1]{\textcolor{agreen}{\textbf{#1}}}
\newcolumntype{Y}{>{\raggedright\arraybackslash}X}
\graphicspath{{./}}

\title{RACF: A Resilient Autonomous Car Framework with Object Distance Correction}

\author{
  \IEEEauthorblockN{Chieh Tsai$^{1}$, Hossein Rastgoftar$^{2}$, Salim Hariri$^{1}$}
  \IEEEauthorblockA{$^{1}$Department of Electrical and Computer Engineering, University of Arizona, Tucson, AZ, USA}
  \IEEEauthorblockA{$^{2}$Department of Aerospace and Mechanical Engineering, University of Arizona, Tucson, AZ, USA}
  \IEEEauthorblockA{vegetableclean@arizona.edu}
}

\begin{document}
\maketitle
\thispagestyle{empty}
\pagestyle{empty}

\begin{abstract}
Autonomous vehicles are increasingly deployed in safety-critical applications, where sensing failures or cyber-physical attacks can lead to unsafe operations resulting in human loss and/or severe physical damages. Reliable real-time perception is therefore critically important for their safe  operations and acceptability. For example, vision-based distance estimation is vulnerable to environmental degradation and adversarial perturbations, and existing defenses are often reactive and too slow to promptly mitigate their impacts on safe operations. We present a Resilient Autonomous Car Framework (RACF) that incorporates an Object Distance Correction Algorithm (ODCA) to improve perception-layer robustness through redundancy and diversity across a depth camera, LiDAR, and physics-based kinematics. Within this framework, when obstacle distance estimation produced by depth camera is inconsistent, a cross-sensor gate activates the correction algorithm to fix the detected inconsistency. We have experiment with the proposed resilient car framework and evaluate its performance on a testbed implemented using the Quanser QCar 2 platform. The presented framework achieved up to 35\% RMSE reduction under strong corruption and improves stop compliance and braking latency, while operating in real time. These results demonstrate a practical and lightweight approach to resilient perception for safety-critical autonomous driving.
\end{abstract}


\afterpage{
\begin{figure*}[t]
  \centering
  \includegraphics[width=0.7\textwidth]{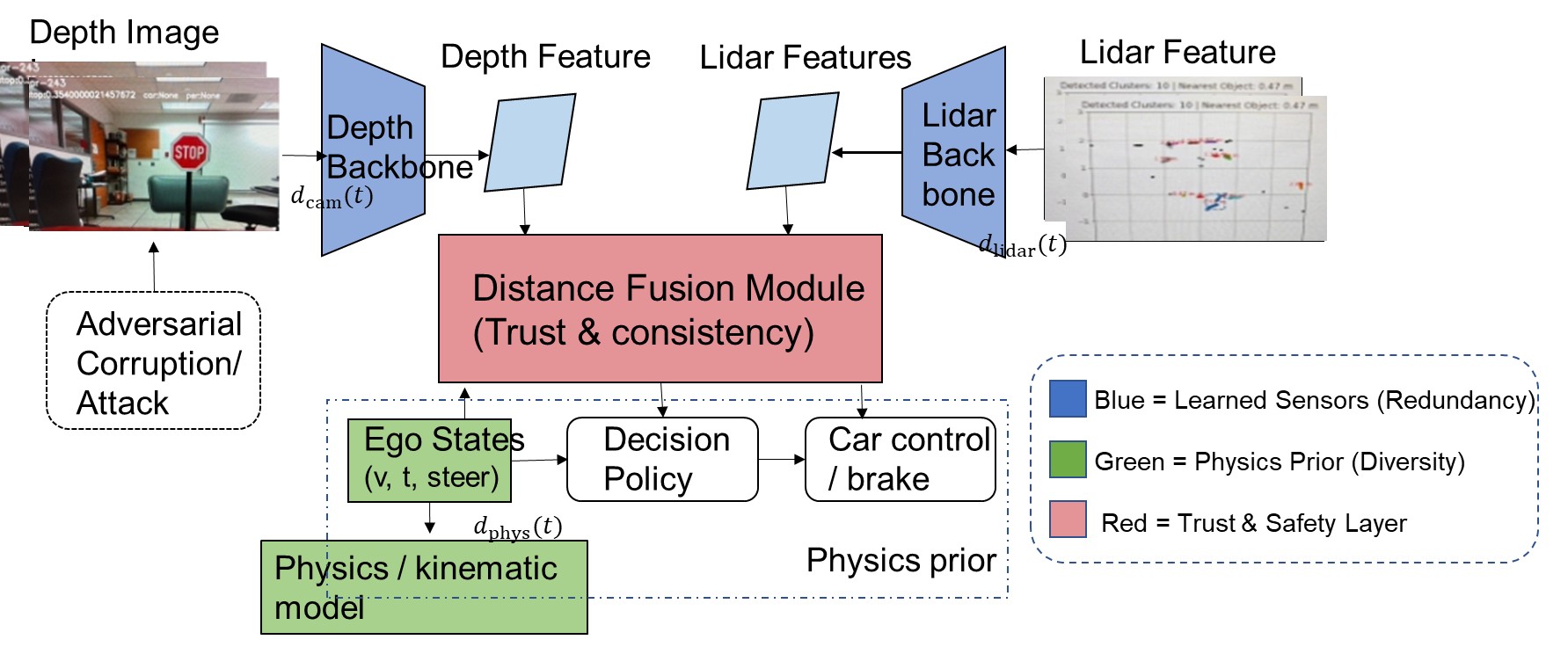}
  \caption{\textbf{Resilient autonomous car framework.}
 Redundancy is achieved via depth and LiDAR sensing, while diversity is provided by a physics/kinematic distance evolution branch. The distance fusion module performs trust \& consistency checks to produce a robust distance estimate for car brake and control functions.}
  \label{fig:resilient_framework}
  \vspace{-2mm}
\end{figure*}
}
\section{Introduction}

Autonomous ground vehicles rely on accurate and timely perception to ensure safe closed-loop operation. Among perception components, obstacle distance estimation plays a critical role in braking and collision avoidance. When control-critical distance signals become corrupted due to adversarial attacks or physical malfunctioning, unsafe behavior such as delayed or missed braking may occur. Vision-based depth sensing, a primary source of obstacle distance estimation, is vulnerable to environmental degradation (e.g., glare and fog)~\cite{zhang2023perception} as well as physical and virtual adversarial perturbations~\cite{zhu2023tpatch}. Such corruption can propagate through the perception--planning--control pipeline and compromise safety~\cite{wong2020targeted}. Maintaining a reliable distance signal under intermittent sensing corruption remains an important practical challenge for real-world autonomous systems.

\begin{figure}[t]
\centering
\includegraphics[width=0.9\columnwidth]{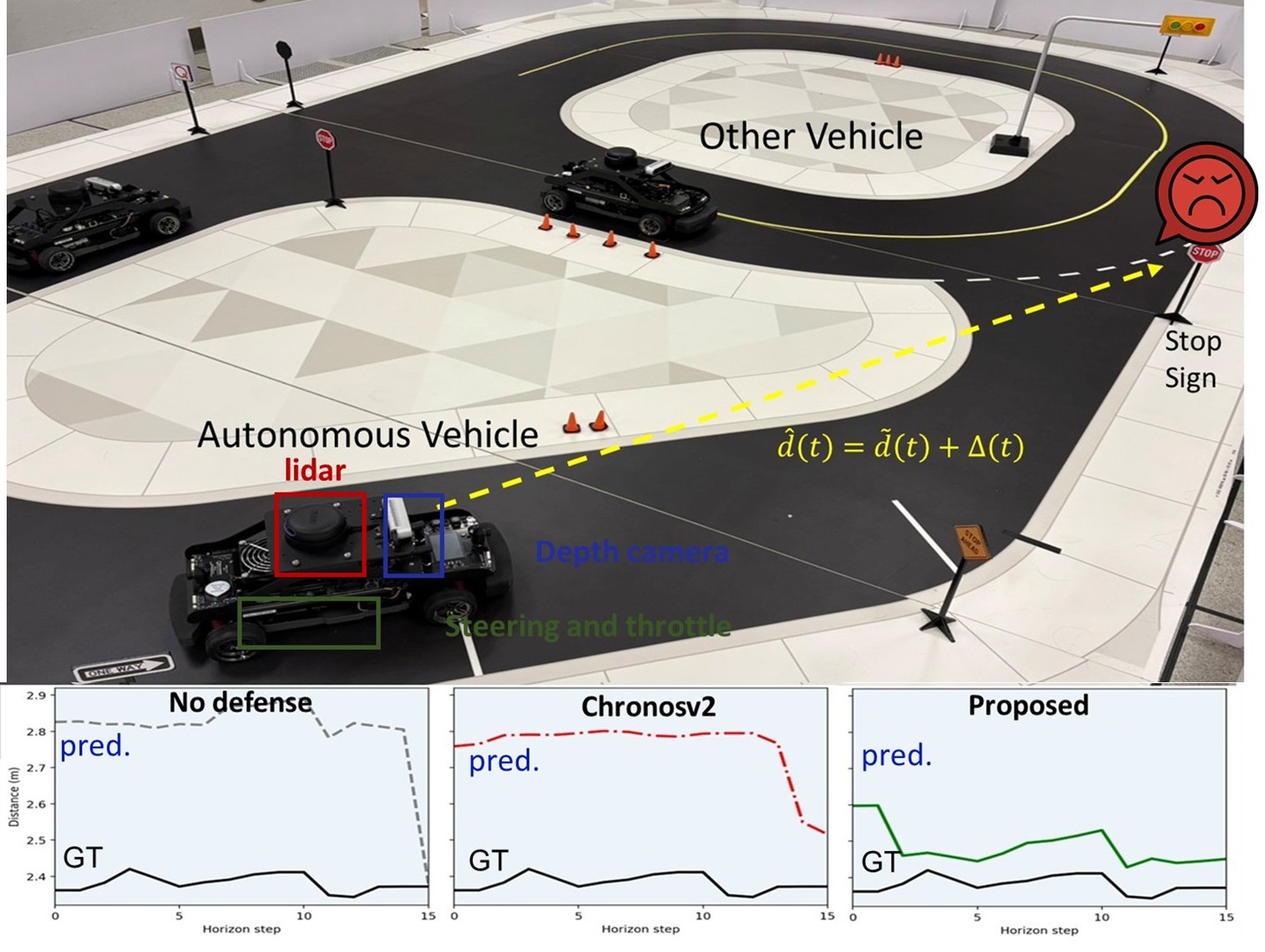}
\caption{
\textbf{Resilient Car Framework Testbed.}
Top: QCar testbed scenario under visual corrup-tion, where attacked depth becomes inconsistent with LiDAR and leads to unsafe distance estimation.
Bottom: Distance predictions for (a) no defense, (b) full replacement using Chronos ~\cite{ansari2025chronos}, and (c) the proposed framework ability to be resilient and robust againt distance sensor attacks.
}
\label{fig:scenario}
\vspace{-2mm}
\end{figure}
Modern platforms typically employ multi-sensor redundancy, for example through camera--LiDAR fusion. However, redundancy alone does not guarantee stable behavior. LiDAR measurements can also suffer from sparsity and noise, and naïve modality switching or unconditional fusion may introduce brittle responses when one modality becomes unreliable~\cite{de2025exploring}. Existing robustness approaches largely focus on input- or model-level defenses, such as adversarial patch mitigation and detector hardening~\cite{zhu2023tpatch,brown2017adversarial,eykholt2018robust,liu2018dpatch}. While these techniques improve detection robustness, they are often reactive and may be too slow to provide reliable real-time safe operation while maintaining nominal performance under clean sensing conditions.

In this paper, we present an architecture to achieve system-level resilient perception. We formulate depth resilience as a selective sensor substitution problem, where corrupted distance measurements due to malicious or natural causes are corrected in real time only when cross-sensor inconsistency is detected. To this end, we introduce a conservative correction mechanism that integrates temporal priors, cross-sensor grounding with LiDAR, and physics-aware kinematic constraints. A residual-based gating strategy activates correction only when depth--LiDAR inconsistency exceeds a predefined threshold. This mechanism mitigates corrupted observations while leaving nominal measurements unchanged.

We implemented the presented resilient car framework on the Quanser QCar~2 platform and evaluated its performance under weak, mid, and strong corruption attack regimes. Offline experiments demonstrate up to 35\% reduction in RMSE under strong attacks without degrading clean-condition performance. Closed-loop stop-sign trials with physical patch injection further show improved stopping compliance and reduced braking latency, illustrating practical real-time resilience in a perception-driven autonomous driving system.

\noindent\textbf{The main contributions can be summarized as follows:}
\begin{itemize}
    \item We formulate depth resilience as a selective signal-level correction problem that preserves nominal measurements while removing intermittent corruptions or maliciously injected distance values.
    
    \item We present the Object Distance Correction Algorithm (ODCA), a conservative residual-based repair mechanism that preserves nominal depth measurements and activates correction only when cross-sensor inconsistency indicates attacks or adversarial manipulation.
    
    \item We develop a multi-sensor evaluation benchmark on the Quanser QCar~2 platform and validate ODCA under both offline corruption and closed-loop physical patch attacks, demonstrating improved distance recovery and braking stability.
    
     \item We leverage a large-scale pretrained foundation time-series model (ChronosV2) and deploy it in real time to provide causal short-horizon temporal priors for conservative distance correction under sensing attacks.
\end{itemize}

\section{Related Work}

\subsection{Autonomous Driving Architecture and Perception Dependencies}

Autonomous driving systems operate through a perception--planning--control pipeline, where reliable sensing is essential for safe operation~\cite{anon_av_resilience_2026}. Multi-sensor configurations, particularly camera and LiDAR combinations, are widely adopted to improve robustness through redundancy. Camera--LiDAR calibration and alignment methods reconcile cross-sensor bias and noise to enable consistent fusion under nominal conditions~\cite{guo2017calibration}, while uncertainty-aware modeling further improves reliability under distribution shift~\cite{lakshminarayanan2017simple}. However, many fusion pipelines implicitly assume reliable sensing and do not explicitly address intermittent corruption of a single modality that can be triggered by adversarial attacks or natural causes.

\subsection{Adversarial and Environmental Attacks on Autonomous Perception}

Vision-based autonomous perception components are vulnerable to both deliberate adversarial manipulation and natural sensing degradation. Physical adversarial patches can mislead traffic-sign and object detectors~\cite{eykholt2018robust,brown2017adversarial}, while adverse environmental conditions such as fog, low illumination, and reflective surfaces introduce systematic perception errors~\cite{lin2014microsoft}. For example, perturbations can propagate to downstream planning and control modules, potentially resulting in unsafe operations.

\subsection{Robustness and Sensor-Level Recovery Approaches}

Most existing robustness techniques focus on image-level defenses or model hardening, aiming to improve visual robustness under attack~\cite{eykholt2018robust,brown2017adversarial}. While effective at improving detection performance, these approaches do not directly correct the control-critical distance signal required for safe navigation planning and control. Multi-sensor fusion provides redundancy across modalities, but naive or unconditional fusion may degrade otherwise clean measurements when one sensor becomes unreliable~\cite{guo2017calibration}. Temporal forecasting offers an alternative perspective by exploiting temporal structure to recover corrupted signals. Recent multivariate forecasting models such as SOFTS~\cite{han2024softs} and foundation time-series models such as Chronos~\cite{ansari2025chronos} demonstrate strong performance across domains. However, most forecasting-based approaches remain forecast-only and lack mechanisms for selectively correcting corrupted measurements while explicitly preserving uncompromised or normal measurements.

\subsection{Research Gap}

Existing research does not provide effective methods to achieve resilient measurements of obstacle distances in real-time under natural or malicious intermittent corruptions. There methods can be briefly described as: (i) image/model-level defenses primarily target visual correctness or detector robustness, but do not explicitly guarantee a stable distance signal for braking; (ii) multi-sensor fusion improves nominal robustness, yet unconditional fusion can inject errors when one modality is unreliable and may also degrade clean measurements; and (iii) forecasting models provide tempo-ral priors, but most are forecast-only and lack conservative mechanisms that repair only when needed with cross-sensor grounding. Our research approach overcomes this gap by developing a selective correction algorithm that combines temporal priors, LiDAR consistency, and physically plausible constraints.
\section{Resilient Autonomous Car Framework}

Resilient autonomous driving requires architectural mechanisms that preserve safety-critical functionality under sensing degradation and adversarial attacks. A robust design leverages both redundancy and diversity across multiple distance estimation sources to ensure correct measurement inputs for planning and control systems~\cite{anon_av_resilience_2026}. Within this architectural view, we consider three complementary methods for obstacle distance estimation: (i) depth camera measurements, (ii) LiDAR-based geometric range estimation, and (iii) physics-based distance estimation derived from vehicle kinematics.

Camera and LiDAR provide cross-modal redundancy, while the physics-based predictor introduces temporal consistency through vehicle dynamics constraints. Rather than blindly fusing these signals, the architecture evaluates cross-sensor consistency and applies conservative selection to correct unreliable measurements. Fig.~\ref{fig:resilient_framework} illustrates our Resilient Autonomous Car Framework (RACF) and highlights the perception-layer component addressed in this paper.In this work, we focus specifically on the perception layer by developing a sensor-level correction mechanism, termed the Object Distance Correction Algorithm (ODCA), that maintains depth camera reliability under intermittent corruption and attack, enabling robust distance estimation for control-critical tasks. In this approach, vehicle kinematics serves as a physics-based consistency constraint within the Object Distance Correction Algorithm (ODCA), stabilizing the correction process without introducing additional control inputs. The control interface continues to operate on a single fused distance signal, preserving the original planning and braking architecture.
\section{Perception-Layer Realization in a Resilient Distance Architecture}
\label{sec:method}
\subsection{Overview}
Figure~\ref{fig:resilient_framework} shows how to apply our resilient autonomous car framework (RACF) to the perception layer, where safe braking critically depends on a trustworthy obstacle distance measurement regardless of sensing degradation and adversarial interference. The architecture combines: (i) redundancy from depth and LiDAR distance evidence, and (ii) diversity from short-horizon physical consistency (kinematics). Rather than unconditional fusion, the perception layer performs cross-sensor validation and outputs a control-ready distance estimate for downstream braking.

\begin{figure}[t]
  \centering
  \includegraphics[width=1.0\linewidth]{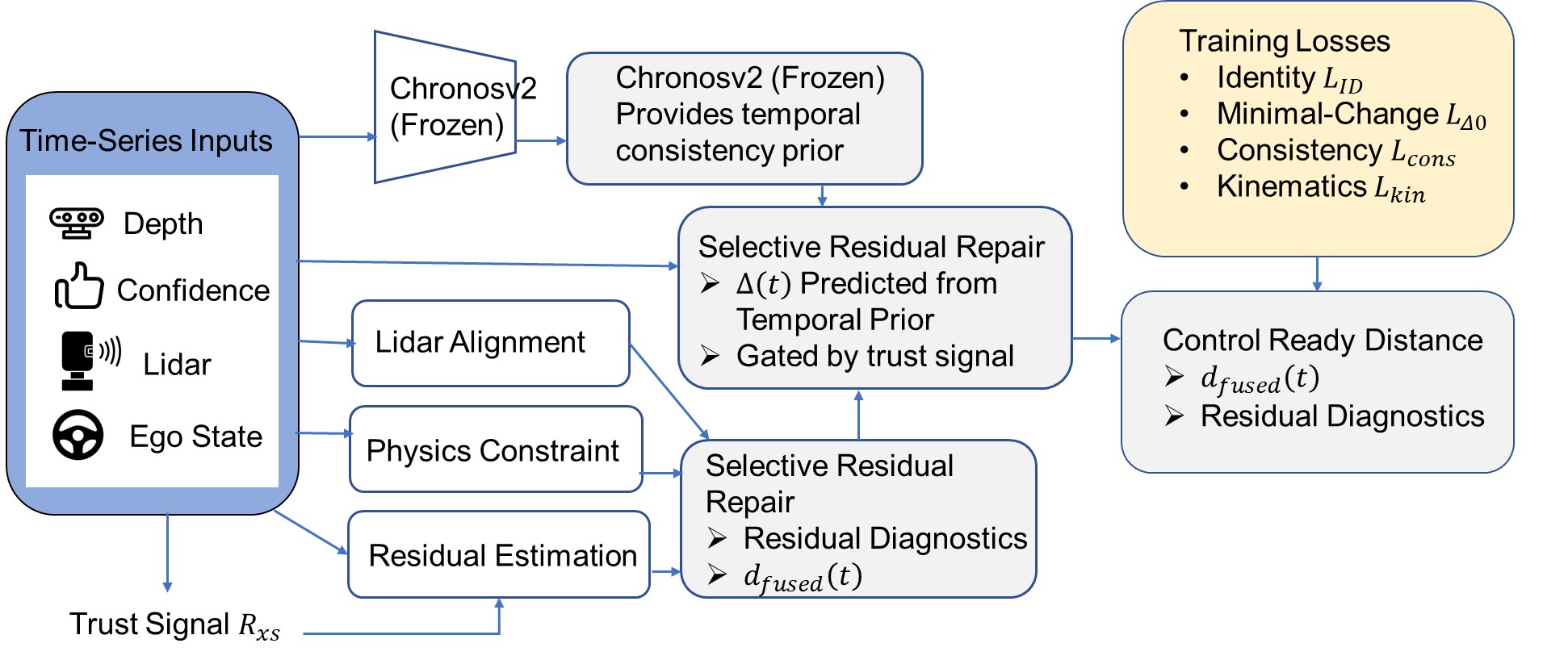}
  \caption{
  \textbf{\textbf{Perception-layer realization of ODCA within RACF.}} ChronosV2 provides uncertainty-aware temporal features. A lightweight delta head predicts additive depth corrections. A LiDAR-based consistency gate controls fusion, producing a conservative control-ready distance output.
  }
  \label{fig:overview}
  \vspace{-2mm}
\end{figure}

We instantiate the Object Distance Correction Algorithm (ODCA), which corrects depth measurements only when necessary and preserves nominal observations otherwise. The ODCA uses: (i) a frozen ChronosV2 temporal prior, (ii) a lightweight additive delta head (921 trainable parameters), and (iii) a LiDAR-driven gate computed from depth--LiDAR inconsistency. Kinematics is used as a training-time regularizer (and optional sanity check), while the deployed gate relies only on cross-sensor inconsistency to avoid high overhead on real-time operations, as shown in Fig.~\ref{fig:overview}.

\subsection{Online Inference}

ODCA is deployed as a causal, per-step correction  module. At each time $t$, it consumes the latest synchronized measurements $(\tilde d(t), \tilde c(t), \ell(t))$ (when LiDAR is available) and outputs a fused distance $\hat d_{\text{fused}}(t)$. Horizon notation is used only for offline training/evaluation.

\begin{algorithm}[t]
\caption{ODCA (Online): Per-Step Gated Depth Repair}
\label{alg:racf}
\begin{algorithmic}[1]
\REQUIRE Context length $W$, Chronos samples $S$
\REQUIRE Depth $\tilde d(t)$, confidence $\tilde c(t)$
\REQUIRE LiDAR $\ell(t)$ (if available)
\ENSURE Fused output $\hat d_{\text{fused}}(t)$

\STATE Form context window $\mathbf{x}_{t-W:t}$ (includes $\tilde d, \tilde c$, and optional states)
\STATE Run frozen ChronosV2 to draw $S$ samples and compute $(\mu(t), \sigma(t))$
\STATE Predict additive correction $\Delta(t)$ using the delta head
\STATE Compute repaired depth: $\hat d_{\text{rep}}(t) \leftarrow \tilde d(t) + \Delta(t)$

\IF{LiDAR available}
    \STATE Compute aligned LiDAR reference $\ell_{\rightarrow D}(t) = \alpha \ell(t) + \beta$
    \STATE Compute residual $r_{\text{XS}}(t) = |\tilde d(t) - \ell_{\rightarrow D}(t)|$
    \STATE Gate 
    $w(t) \leftarrow 
    \mathrm{clip}\!\left(
    \frac{r_{\text{XS}}(t) - \tau_{\text{low}}}
         {\tau_{\text{high}} - \tau_{\text{low}}},
    0, 1
    \right)^{\gamma}$
\ELSE
    \STATE $w(t) \leftarrow 1$ \quad (fallback to repair-only)
\ENDIF

\STATE Fuse conservatively:
$\hat d_{\text{fused}}(t) \leftarrow (1 - w(t))\,\tilde d(t) + w(t)\,\hat d_{\text{rep}}(t)$

\RETURN $\hat d_{\text{fused}}(t)$
\end{algorithmic}
\end{algorithm}

\subsection{ChronosV2 Temporal Featurization}

ChronosV2 is treated as a frozen probabilistic forecaster. Given context length $W$, it produces $S$ forecast samples whose mean and standard deviation form uncertainty-aware temporal features. We build the delta-head input by concatenating the latest observation features (e.g., $\tilde d(t),\tilde c(t)$) with $(\mu(t),\sigma(t))$ (and vehicle states when available). The sampling period $\Delta t$ is estimated from timestamps (median inter-sample interval) or set to a fixed value (e.g., $0.02$\,s at 50\,Hz).

\subsection{Delta Repair Head}

The delta head predicts an additive correction:
\[
\hat d_{\text{rep}}(t)=\tilde d(t)+\Delta(t).
\]
This residual formulation naturally supports \emph{selective repair}: under nominal sensing, $\Delta(t)$ is driven toward zero and the final output remains close to the original measurement.

\subsection{Cross-Sensor Alignment and Gate}

To compare depth and LiDAR in a common domain, we use an affine alignment:
\[
\ell_{\rightarrow D}(t) = \alpha \ell(t) + \beta ,
\]
where $\ell_{\rightarrow D}(t)$ denotes the LiDAR range aligned to the depth-distance domain.
The alignment parameters $(\alpha,\beta)$ are estimated via robust least-squares over a short window using high-confidence depth samples (or an initial clean calibration run) and kept fixed within a run.

The gate is computed from depth--LiDAR inconsistency:
\[
w(t)=\mathrm{clip}\!\left(
\frac{|\tilde d(t)-\ell_{\rightarrow D}(t)|-\tau_{\text{low}}}
{\tau_{\text{high}}-\tau_{\text{low}}},
0,1
\right)^{\gamma}.
\]

When sensors agree, $w(t)\!\approx\!0$ and $\hat d_{\text{fused}}(t)\!\approx\!\tilde d(t)$; 
when disagreement grows, $w(t)$ increases and the output relies more on the repaired signal.

In practice, we observed that direct regression between LiDAR range and camera-based depth estimates is highly unstable due to differences in sensing geometry and object localization noise. Empirically, naive regression produced large residual errors and unstable distance estimates in our dataset. Therefore, LiDAR is used only as a consistency trigger rather than as a direct substitute for the depth measurement.

\noindent\textbf{Proposition 1 (Nominal preservation under agreement).}
Assume LiDAR is available and the gate is defined as
\(
w(t)=\mathrm{clip}\!\left(\frac{r_{\mathrm{XS}}(t)-\tau_{\mathrm{low}}}{\tau_{\mathrm{high}}-\tau_{\mathrm{low}}},0,1\right)^{\gamma}
\),
where \(r_{\mathrm{XS}}(t)=|\tilde d(t)-\ell_{\rightarrow D}(t)|\).
If \(r_{\mathrm{XS}}(t)\le \tau_{\mathrm{low}}\), then \(w(t)=0\) and the fused output satisfies
\[
\hat d_{\mathrm{fused}}(t)=\tilde d(t),
\]
i.e., the measurement is passed through unchanged.
Moreover, for any \(t\), the fusion is a convex combination:
\[
\hat d_{\mathrm{fused}}(t)=(1-w(t))\,\tilde d(t)+w(t)\,\hat d_{\mathrm{rep}}(t),\quad w(t)\in[0,1],
\]
so the output continuously interpolates between nominal sensing and repaired sensing as cross-sensor disagreement grows.

\subsection{Training Objectives}

Training follows a corruption-aware objective that preserves clean measurements while correcting corrupted segments:
(i) \textbf{identity preservation} on clean/high-confidence steps,
(ii) \textbf{minimal-change} regularization to discourage unnecessary correction,
(iii) \textbf{cross-sensor consistency} to keep repaired depth aligned with $\ell_{\rightarrow D}$ when disagreement is high, and
(iv) \textbf{kinematics regularization} applied on first differences of the repaired signal to encourage short-horizon physical plausibility (e.g., $\Delta d \approx -v\Delta t$).
The total loss is:
\[
\mathcal{L}=
\lambda_{\text{ID}}\mathcal{L}_{\text{ID}}
+\lambda_{\Delta 0}\mathcal{L}_{\Delta 0}
+\lambda_{\text{cons}}\mathcal{L}_{\text{cons}}
+\lambda_{\text{kin}}\mathcal{L}_{\text{kin}},
\]
with larger weights applied to low-confidence or attacked regions.

\subsection{Residual-Based Diagnostic Signals}

ODCA exposes interpretable residuals for diagnosis:
\begin{equation}
\begin{aligned}
r_{\text{XS}}(t) &= \left| \tilde d(t) - \ell_{\rightarrow D}(t) \right|, \\
r_{\Delta}(t) &= \left| \Delta(t) \right|, \\
r_{\text{post}}(t) &= \left| \hat d_{\text{fused}}(t) - \ell_{\rightarrow D}(t) \right|.
\end{aligned}
\end{equation}

These residuals support attack diagnostic evaluation (AUROC/AUPRC) and complement recovery accuracy.

\section{Experimental Setup}
\label{sec:setup}

\subsection{Overview}

Vision-based depth estimation is vulnerable to environmental and adversarial corruption, potentially leading to unsafe braking. We therefore adopt a controlled attack-evaluation 

\subsection{QCar~2 Multi-Sensor Driving Benchmark}

We collect synchronized multi-sensor data on a Quanser QCar~2 platform using the official indoor map at $\approx$50\,Hz, totaling 67{,}821 samples across 13 sequences under speeds $\{1.0, 1.5, 2.0\}$\,m/s and steering angles $\{0^\circ, \pm15^\circ\}$.

Each timestamp includes depth-based distance and confidence, 2D LiDAR ranges, and vehicle state variables. The clean indoor depth signal is treated as pseudo-ground truth. All reported RMSE/MAE values are evaluated relative to this clean-depth baseline under the same sensing pipeline.

Stop signs are localized via YOLOv8-seg~\cite{yaseen2024yolov8}, and obstacle distance is computed as the median depth within the segmentation mask. LiDAR returns are clustered using DBSCAN~\cite{ester1996dbscan}, where the nearest cluster provides a cross-sensor reference.

\subsection{Hybrid Attack Evaluation}

\begin{figure}[t]
\centering
\begin{subfigure}[t]{0.51\columnwidth}
  \centering
  \includegraphics[width=\linewidth]{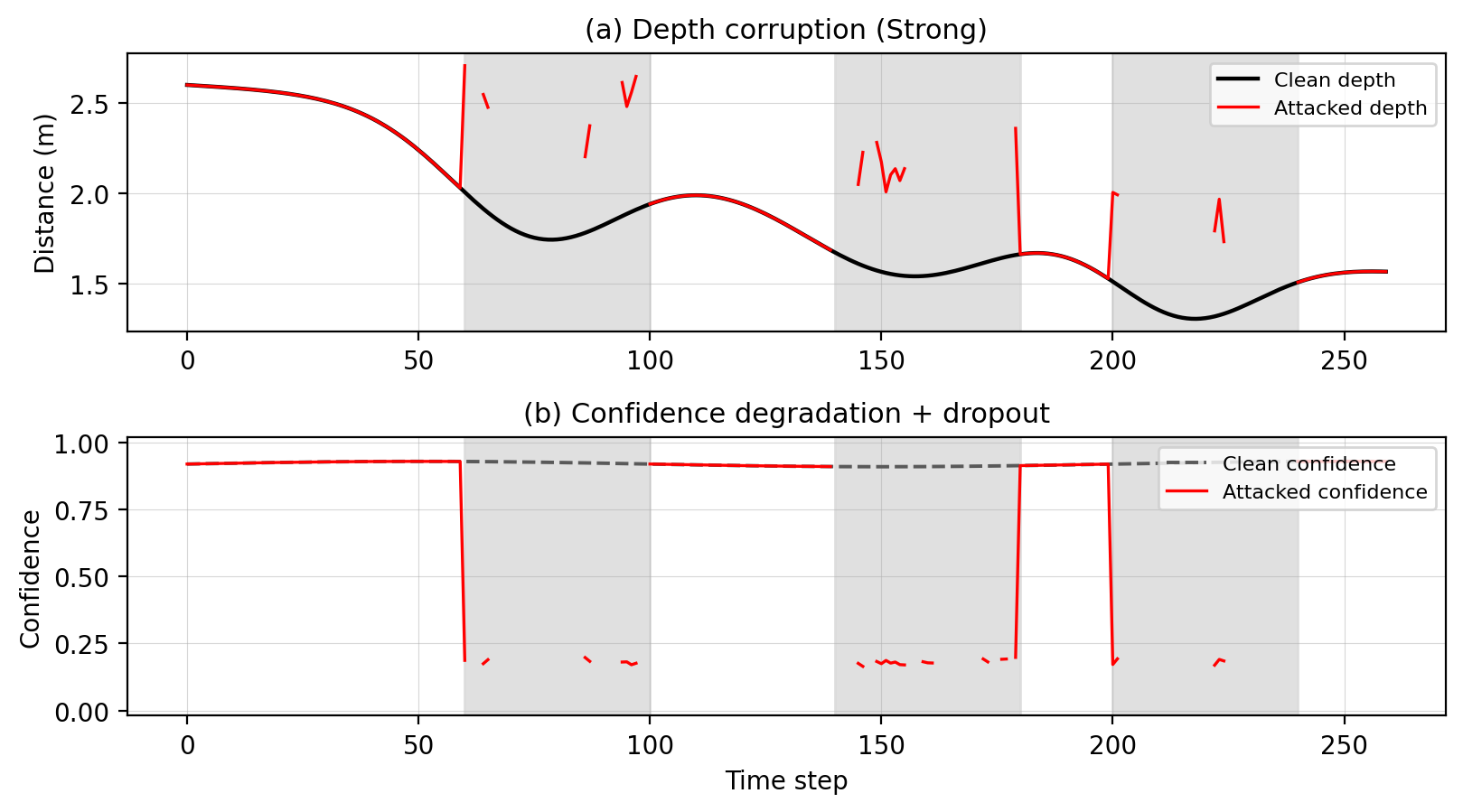}
  \caption{Offline signal corruption (strong).}
\end{subfigure}\hfill
\begin{subfigure}[t]{0.39\columnwidth}
  \centering
  \includegraphics[width=\linewidth]{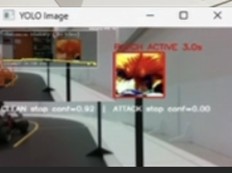}
  \caption{Online adversarial patch suppressing stop-sign detection.}
\end{subfigure}

\vspace{-1mm}
\caption{\textbf{Attack scenarios used in evaluation.}
(a) Controlled offline corruption applied to depth/confidence signals.
(b) Physical adversarial patch deployed during live operation, suppressing stop-sign detection and inducing distance dropout.}
\label{fig:attack_overview}
\vspace{-2mm}
\end{figure}

We evaluate resilience under both offline corruption and closed-loop physical patch attacks. Offline attacks introduce depth bias, confidence degradation, and blackout segments at three severity levels (weak/mid/strong), enabling controlled benchmarking against a clean reference (Fig.~\ref{fig:attack_overview}(a)).

For real-world validation, we conduct closed-loop patch trials following~\cite{hu2021naturalistic}. Attack episodes are triggered after stable stop-sign detection and remain active during the critical braking window, with a short cooldown to prevent retriggering (Fig.~\ref{fig:attack_overview}(b)). Per-frame detection status, estimated distance, brake commands, and timestamps are logged.

A trial is considered successful only if braking is triggered and the stopping distance satisfies a predefined safety threshold; late braking is treated as failure. Since resilience depends on the relation between attack duration $T_{\text{atk}}$ and system reaction time $T_{\text{react}}$, we report braking latency $\Delta t_{\text{brake}}$ as a control-relevant metric and analyze suppression behavior under varying attack durations (Table~\ref{tab:attack_persistence}).

\begin{table}[t]
\centering
\caption{Attack persistence analysis under varying patch durations (no defense).}
\label{tab:attack_persistence}
\footnotesize
\setlength{\tabcolsep}{6pt}
\begin{tabular}{c|cc}
\toprule
$T_{\text{atk}}$ (s) 
& Lost-Detection Frames 
& ASR $\uparrow$ \\
\midrule
0.5  & 7--8   & 0.50 \\
1.0  & 15--16 & 0.90 \\
3.0  & 48--58 & 1.00 \\
\bottomrule
\end{tabular}
\end{table}
\subsection{Problem Definition}
Given corrupted depth $\tilde d(t)$, confidence $\tilde c(t)$, and LiDAR references $\ell(t)$ when available, our goal is to output a repaired distance estimate $\hat d(t)$ that is accurate and temporally stable under intermittent corruption.
Performance is evaluated against a clean depth baseline (pseudo-ground truth) and cross-sensor consistency metrics. The clean baseline is used only for offline evaluation and is not available at inference time.

\section{Experimental Results}
\label{sec:results}

\subsection{Evaluation Metrics}

We evaluate performance from two perspectives: \textbf{recovery fidelity} and \textbf{diagnostic utility}.
Recovery fidelity is measured by \textbf{MAE} and \textbf{RMSE} ($\downarrow$) between the fused estimate $\hat d_{\text{fused}}$ and a clean reference.
Diagnostic quality is assessed using \textbf{AUROC} and \textbf{AUPRC} ($\uparrow$), computed from the cross-sensor residual 
$r_{\text{XS}}(t)=|\tilde d(t)-\ell_{\rightarrow D}(t)|$ (\textbf{XS}) 
and the correction magnitude 
$r_{\text{chg}}(t)=|\hat d_{\text{fused}}(t)-\tilde d(t)|$ (\textbf{Chg}).

Beyond pointwise error, resilience is reflected in bounded degradation across weak/mid/strong attacks, conservative correction that preserves nominal behavior, and control relevance verified by stable closed-loop braking.

To summarize robustness across attack strengths, we report 
\textbf{Bounded Degradation (BD)}:
$\mathrm{BD}=(\mathrm{RMSE}_{\mathrm{strong}}-\mathrm{RMSE}_{\mathrm{weak}})/\mathrm{RMSE}_{\mathrm{weak}}$ (lower is better),
and \textbf{Resilience Gain Ratio (RGR)}:
$\mathrm{RGR}(s)=1-\mathrm{RMSE}_{\mathrm{ours}}(s)/\mathrm{RMSE}_{\mathrm{Chronos}}(s)$ (higher is better),
averaged over $s\in\{\mathrm{weak,mid,strong}\}$.

For real-world trials, we additionally measure \textbf{braking latency} $\Delta t_{\text{brake}}$, defined as the time from the first valid stop trigger to reaching the stop threshold $d_{\text{brake}}\le 0.60$\,m.

\begin{figure}[t]
\centering
\includegraphics[width=\columnwidth]{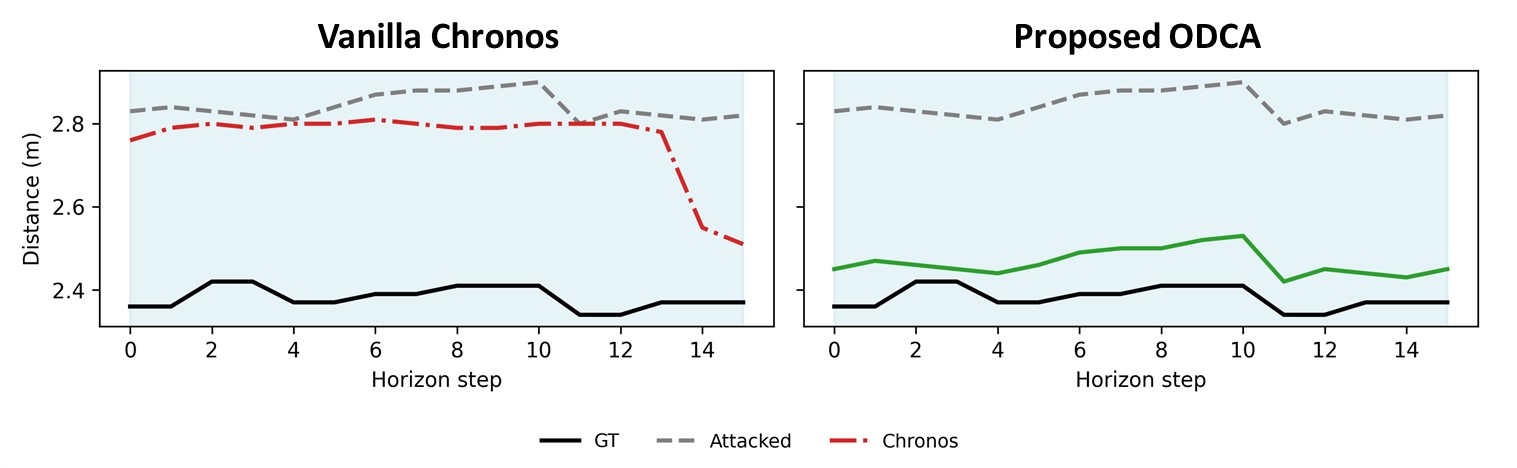}
\caption{
Strong-strength attack comparison.
Left: Chronos~\cite{ansari2025chronos}.
Right: ODCA (Ours).
Black: clean-depth reference (pseudo-ground truth); gray dashed: attacked input.
ODCA preserves nominal segments and selectively corrects corrupted intervals.
}
\label{fig:qual_compare}
\vspace{-2mm}
\end{figure}

\begin{table}[t]
\centering
\caption{Distance estimation accuracy under different attack strengths.
Lower RMSE / MAE indicate better perception recovery ($\downarrow$).}
\label{tab:recovery}
\scriptsize
\renewcommand{\arraystretch}{1.15}
\setlength{\tabcolsep}{4pt}
\begin{tabular}{l|cc|cc|cc}
\toprule
& \multicolumn{6}{c}{\textbf{Attack Strength}} \\
\cmidrule(lr){2-7}
\textbf{Model}
& \multicolumn{2}{c}{Weak}
& \multicolumn{2}{c}{Mid}
& \multicolumn{2}{c}{Strong} \\
\cmidrule(lr){2-3} \cmidrule(lr){4-5} \cmidrule(lr){6-7}
& RMSE & MAE & RMSE & MAE & RMSE & MAE \\
\midrule

\rowcolor{rowgray}
\textbf{ODCA (Ours)}
& \textcolor{bestblue}{\textbf{0.111}} & \textcolor{bestblue}{\textbf{0.093}}
& \textcolor{bestblue}{\textbf{0.193}} & \textcolor{bestblue}{\textbf{0.170}}
& \textcolor{bestblue}{\textbf{0.323}} & \textcolor{bestblue}{\textbf{0.283}} \\

Chronos~\cite{ansari2025chronos}
& 0.229 & 0.151
& 0.363 & 0.256
& 0.503 & 0.353 \\

SOFTS~\cite{han2024softs}
& 0.207 & 0.171
& 0.330 & 0.232
& 0.649 & 0.487 \\

NHITS~\cite{challu2023nhits}
& 0.144 & 0.104
& 0.300 & 0.214
& 0.498 & 0.365 \\

DLinear~\cite{zeng2023transformers}
& 0.153 & 0.117
& 0.315 & 0.223
& 0.657 & 0.530 \\

NLinear~\cite{zeng2023transformers}
& 0.157 & 0.116
& 0.363 & 0.261
& 0.708 & 0.577 \\
\bottomrule
\end{tabular}
\end{table}

\begin{figure}[t]
  \centering
  \includegraphics[width=0.6\columnwidth]{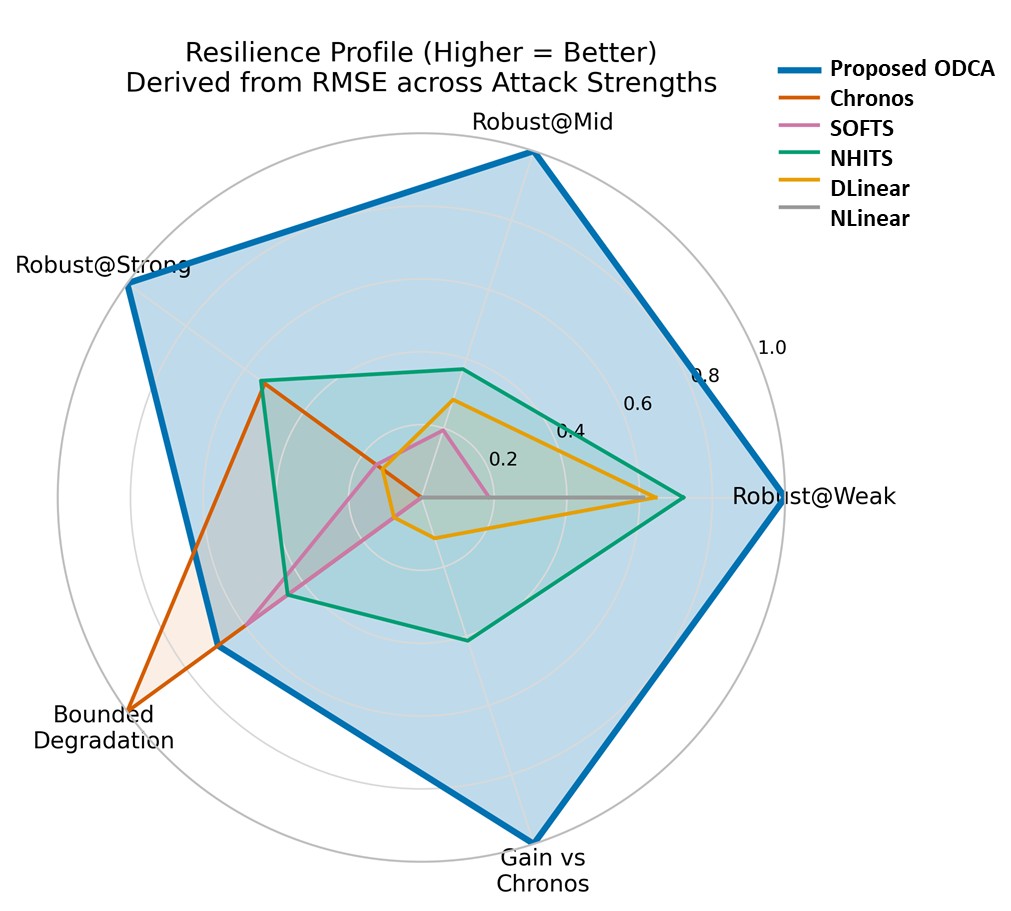}
  \caption{\textbf{Resilience summary across attack strengths (higher is better).}
  Radar axes aggregate RMSE-based robustness at weak/mid/strong, bounded degradation (BD), and average gain vs.\ Chronos (RGR).}
  \label{fig:resilience_radar}
  \vspace{-2mm}
\end{figure}

\begin{figure*}[!t]
  \centering
  \includegraphics[width=0.96\textwidth]{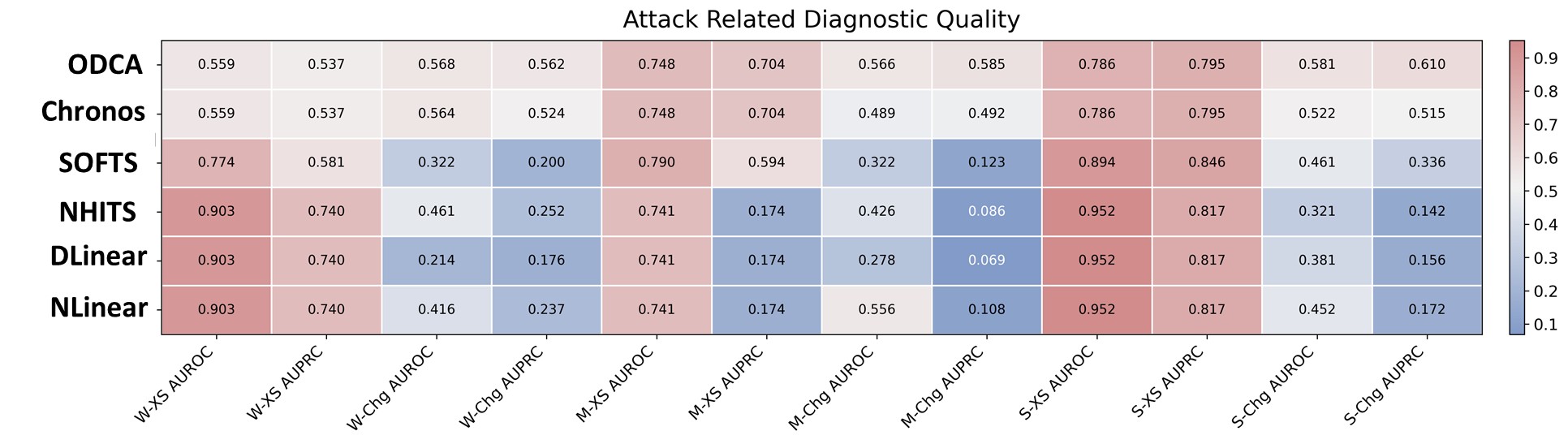}
  \caption{
  \textbf{Attack-related diagnostic quality ($\uparrow$ better).}
  Heatmap visualization of AUROC and AUPRC across weak, mid, and strong attack regimes.
  Cross-sensor residual (XS) and correction magnitude (Chg) are reported for each method.
  Exact numerical values are overlaid for clarity.
  }
  \label{fig:diag_heatmap}
  \vspace{-2mm}
\end{figure*}
\definecolor{secondgray}{gray}{0.35}
\newcommand{\bestb}[1]{\textcolor{bestblue}{\textbf{#1}}}   
\newcommand{\bestg}[1]{\textcolor{secondgray}{\textbf{#1}}} 
\begin{figure}[t]
\centering
\includegraphics[width=0.65\columnwidth]{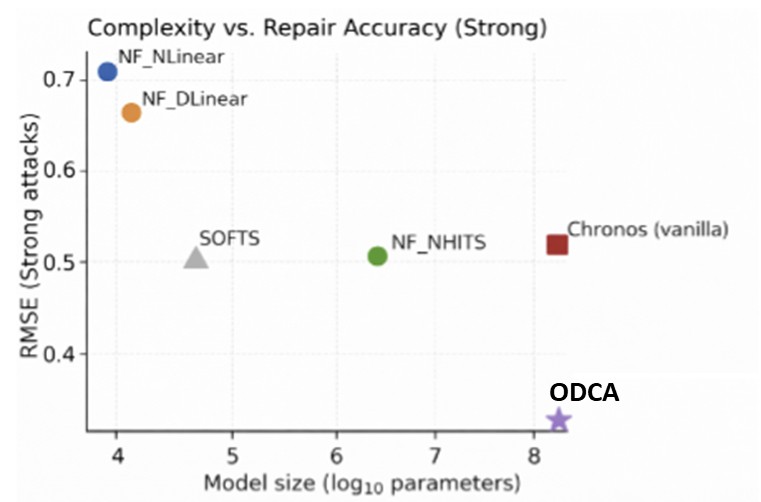}
\caption{Model complexity vs.\ repair accuracy (Strong). Chronos is frozen; ODCA adds 921 trainable parameters.}
\label{fig:complexity}
\vspace{-2mm}
\end{figure}

\subsection{Experimental Comparison and Results}

We compare ODCA with representative forecasting baselines (citations embedded in Table~\ref{tab:recovery} and Fig.~\ref{fig:diag_heatmap}). A by-series split is adopted (70\% train, 10\% validation, 20\% test). ChronosV2 remains frozen and only the 921-parameter delta head is trained. The prediction horizon is fixed at $H=16$, corresponding to $\approx 0.32$\,s at 50\,Hz, aligning with near-term braking decisions.
Quantitative results are summarized in Table~\ref{tab:recovery} (RMSE/MAE) and Fig.~\ref{fig:diag_heatmap}; XS residuals exhibit strong separability across models, while ODCA demonstrates consistently higher Chg-AUPRC under mid and strong attacks. Fig.~\ref{fig:qual_compare} provides qualitative comparison under strong corruption, illustrating selective correction that preserves nominal segments. In terms of complexity (Fig.~\ref{fig:complexity}), ODCA maintains essentially the same parameter scale as Chronos since the backbone is frozen; performance gains arise from cross-sensor validation and gated repair rather than increased model capacity. Fig.~\ref{fig:resilience_radar} summarizes robustness across weak/mid/strong attacks, bounded degradation (BD), and resilience gain relative to Chronos (RGR). Larger area indicates stronger robustness across regimes (BD is inverted so that higher is better).

\paragraph{Comparison with classical sensor baselines.}

To contextualize the proposed framework, we compare ODCA with
representative distance estimation approaches, including direct
LiDAR measurements, EKF-based temporal filtering, and Chronos
forecasting. These baselines provide reference points for
traditional sensing, filtering, and forecasting pipelines rather
than direct competitors to the proposed corruption-aware
correction mechanism. For each baseline, the distance estimate is obtained directly from
the corresponding sensor or filter output, and RMSE is computed
with respect to the same clean-depth reference used as pseudo-ground
truth in our evaluation protocol. This ensures a consistent
comparison across methods under identical corrupted input sequences. While EKF fusion can reduce measurement noise through temporal
filtering, it assumes approximately Gaussian noise and does not
explicitly address intermittent or adversarial sensor corruption.
The relatively large error of LiDAR measurements is mainly due to
the sparse returns and limited angular resolution of the 2D LiDAR
used on our platform.

\begin{table}[t]
\centering
\caption{Comparison with representative sensor-level baselines under strong attacks.}
\label{tab:sensor_baseline}
\footnotesize
\setlength{\tabcolsep}{6pt}
\begin{tabular}{lc}
\toprule
Method & RMSE $\downarrow$ \\
\midrule
LiDAR measurement & 0.600 \\
EKF fusion~\cite{fujii2013extended} & 0.450 \\
Chronos forecast~\cite{ansari2025chronos} & 0.503 \\
\rowcolor{rowgray}
ODCA (Ours) & \textbf{0.323} \\
\bottomrule
\end{tabular}
\end{table}
\begin{figure}[t]
  \centering
\includegraphics[width=\columnwidth]{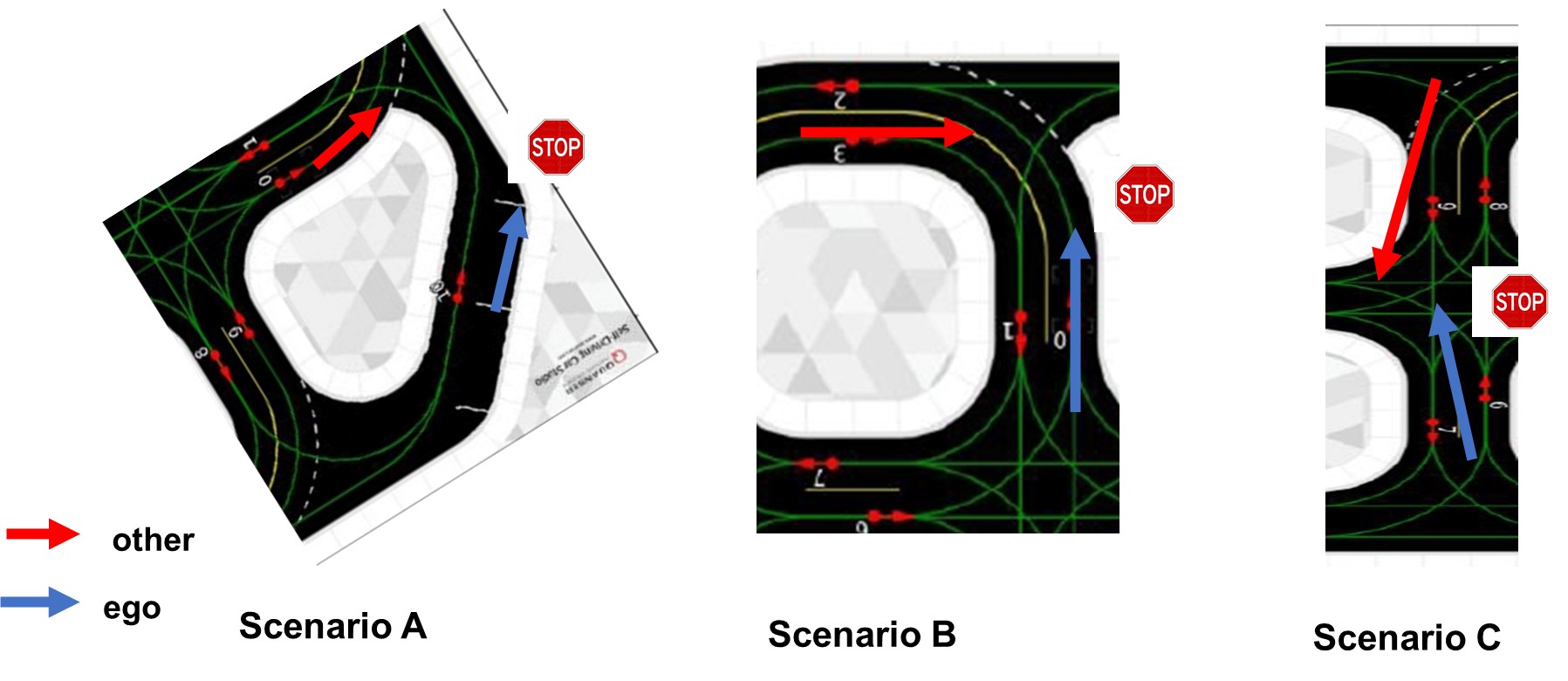}
  \caption{\textbf{High-risk stop-sign scenarios on the indoor map.}
  The ego vehicle (blue arrow) encounters cross-traffic from other vehicles (red arrow) at conflict-prone intersections. 
  The stop sign is placed in these regions to evaluate braking safety under attack.}
  \label{fig:stop_sign_scenarios}
  \vspace{-2mm}
\end{figure}
\subsection{Real-World Implementation}
\label{subsec:realworld}

We deploy QCar~2 on the official indoor map with multiple turns and intersection-like decision points (Fig.~\ref{fig:stop_sign_scenarios}). A YOLO-based detector localizes the stop sign online, and braking is triggered based on accumulated detections and the estimated stop-sign distance

\paragraph{Closed-loop metrics.}
We report Stop Compliance Rate (SCR) and Attack Success Rate (ASR) for closed-loop trials, along with braking latency $\Delta t_{\text{brake}}$.

\paragraph{Patch trials and integration.}
Following the physical patch protocol in~\cite{hu2021naturalistic}, we overlay an adversarial patch on the detected stop-sign region for a randomly sampled duration $T_{\text{atk}}\sim\mathcal{U}(0.5,3.0)$\,s, aligned to the critical braking phase. To ensure consistent episode alignment across methods, an attack episode is triggered only after a stable stop-sign detection is observed, and a short cooldown prevents immediate retriggering. We log per-frame detection status and confidence, estimated stop-sign distance, brake command, and timestamps.

\paragraph{Patch coverage sensitivity.}
In preliminary closed-loop sweeps, we found that attack success is dominated by the patch occlusion ratio $\rho \in [0,1]$ over the stop-sign region. With partial coverage ($\rho \in \{0.3, 0.6\}$), detector confidence decreases but intermittent re-detections still occur, which remains sufficient for the controller to accumulate detections and trigger braking (ASR = 0). In contrast, near-full coverage ($\rho \in \{0.9, 1.0\}$) consistently suppresses detection throughout the braking window, resulting in missed or late braking beyond the safety margin (ASR = 1.0). Therefore, we treat $\rho \in [0.9, 1.0]$ as the effective strong physical attack regime in Table~\ref{tab:realworld}.

\paragraph{Offboard deployment and evaluation.}
Due to limited onboard compute, ODCA runs on an offboard server over a local network. 
Perception streams are transmitted at $\approx 50$\,Hz and brake decisions are returned to the vehicle; 
end-to-end latency is recorded to verify real-time feasibility. 
From system logs, the average processing cycle is approximately 49\,ms per frame ($\approx$20\,Hz), 
including perception transmission, ODCA inference, and control feedback. 
This confirms that the proposed framework operates within the real-time constraints of the QCar~2 platform 
while introducing only 921 additional trainable parameters. 
We conduct 10 closed-loop trials per setting (Clean, Attack, Attack+ODCA), with two stop encounters per run. A stop is counted as successful only if braking is triggered and $d_{\text{brake}} \le 0.60$\,m; 
late braking is treated as failure even if the vehicle eventually stops.
\paragraph{Statistical details.}
SCR is computed as $\mathrm{SCR}=N_{\text{success}}/N$ with $N=30$ stops per setting,
and ASR is $1-\mathrm{SCR}$ under attack settings.
Braking latency $\Delta t_{\text{brake}}$ is reported as mean $\pm$ standard deviation
over successful stops only, together with the sample size $N_{\Delta t}$.

\begin{table}[t]
\centering
\caption{Closed-loop stop-sign trials on QCar~2 (real-world), $N=30$ per setting.
SCR and ASR are trial-level rates; $\Delta t_{\text{brake}}$ is computed over successful stops only.}
\label{tab:realworld}
\footnotesize
\setlength{\tabcolsep}{4pt}
\begin{tabular}{l|ccc}
\toprule
\textbf{Setting} 
& \textbf{SCR} $\uparrow$ 
& \textbf{ASR} $\downarrow$ 
& $\Delta t_{\text{brake}}$ (s) $\downarrow$ \\
\midrule

Clean (no attack) 
& 1.00 (30/30) 
& 0.00 (0/30) 
& $0.1647 \pm 0.0160$ \\

Attack (patch~\cite{hu2021naturalistic}, no defense) 
& \textbf{0.17 (5/30)} 
& \textbf{0.83 (25/30)} 
& $2.65 \pm 0.21$ \\

Attack + ODCA (patch~\cite{hu2021naturalistic}) 
& 0.80 (24/30) 
& 0.20 (6/30) 
& $0.1900 \pm 0.0300$ \\

\bottomrule
\end{tabular}
\end{table}

\subsection{Ablation on Training Losses}
\label{sec:ablation}

To validate the necessity of each loss component, we evaluate incremental training objectives under the \textbf{Strong} attack setting on the same held-out split as the main experiments. We report repair fidelity (RMSE) and diagnostic utility of correction magnitude (AUROC) as shown in Table~\ref{tab:ablation_strong} 
\begin{table}[t]
\centering
\caption{Ablation under \textbf{Strong} attacks.
Lower RMSE is better ($\downarrow$); higher AUROC is better ($\uparrow$).
Full objective achieves the best recovery while maintaining diagnostic interpretability.}
\label{tab:ablation_strong}

\vspace{-1mm}
\scriptsize
\renewcommand{\arraystretch}{1.08}
\setlength{\tabcolsep}{4pt}

\begin{tabular}{lccc}
\toprule
\textbf{Training Objective}
& \textbf{RMSE$_{\text{rep}}\downarrow$}
& $\Delta$RMSE
& \textbf{AUROC$_{\text{chg}}\uparrow$} \\
\midrule

$\mathcal{L}_{\mathrm{ID}}$
& 0.588
& -- 
& 0.425 \\

$\mathcal{L}_{\mathrm{ID}} + \mathcal{L}_{\Delta 0}$
& 0.393
& -0.195
& 0.609 \\

$\mathcal{L}_{\mathrm{ID}} + \mathcal{L}_{\Delta 0} + \mathcal{L}_{\mathrm{cons}}$
& 0.366
& -0.222
& \bestAUROC{0.648} \\

$\mathcal{L}_{\mathrm{ID}} + \mathcal{L}_{\Delta 0} + \mathcal{L}_{\mathrm{kin}}$
& 0.374
& -0.214
& 0.593 \\

\rowcolor{alight}
$\mathcal{L}_{\mathrm{ID}} + \mathcal{L}_{\Delta 0} + \mathcal{L}_{\mathrm{cons}} + \mathcal{L}_{\mathrm{kin}}$ (Full)
& \bestRMSE{0.362}
& -0.226
& 0.576 \\

\bottomrule
\end{tabular}

\vspace{2pt}

\scriptsize
\textit{Observation:}
$\mathcal{L}_{\Delta 0}$ yields the largest immediate recovery gain.
$\mathcal{L}_{\mathrm{cons}}$ improves diagnostic separability (AUROC),
while $\mathcal{L}_{\mathrm{kin}}$ stabilizes dynamics and yields the best overall RMSE.
\end{table}
\subsection{Analysis and Discussion}

ODCA achieves the lowest RMSE and MAE across all attack levels (Table~\ref{tab:recovery}), demonstrating robust distance recovery under increasing corruption. While several baselines obtain high XS-AUROC/AUPRC (Fig.~\ref{fig:diag_heatmap}), indicating that cross-sensor residual rXS is an effective inconsistency cue, detection alone does not ensure accurate reconstruction. ODCA applies gated correction only under depth–LiDAR disagreement, im-proving both recovery accuracy and Change-AUROC/AUPRC.

LiDAR is not used as a direct substitute; instead, it acts as a consistency trigger, preserving nominal depth under agreement and activating repair only when necessary. Together with identity, consistency, and kinematic regularization ($\Delta d \approx -v\Delta t$), this yields stable and physically plausible recovery.

In closed-loop trials (Table~\ref{tab:realworld}), performance depends on visibility timing and reaction window. Residual failures (SCR $=0.80$) occur mainly under the 3\,s continuous attack regime, where attack persistence exceeds the effective causal recovery horizon. This reflects a reaction-time bound rather than instability of the repair mechanism. The current prototype operates offboard, and end-to-end latency may further reduce the effective repair window in short or turning approaches. Future work will integrate embedded deployment and explore complementary supervisory safety mechanisms to strengthen resilience under sustained attacks.

\section{Conclusion}

We studied perception-layer resilience for autonomous driving under intermittent depth corruption and proposed \textbf{ODCA}, a conservative sensor substitution mechanism that preserves a control-critical distance signal for safe braking. Our approach leverages a pretrained time-series foundation model as a frozen temporal prior, capturing generic short-horizon dynamics without task-specific retraining. By learning only a lightweight additive repair head on top of this foundation prior, ODCA enables correction-only-when-needed behavior through a depth--LiDAR inconsistency gate, while preserving nominal measurements under agreement. Experiments on the Quanser QCar~2 platform demonstrate consistent improvements over forecast-only baselines without degrading clean-condition performance. Closed-loop stop-sign trials with physical patch injection further validate improved stop compliance and reduced braking latency. Beyond ground vehicles, the conservative cross-sensor substitution principle grounded in foundation-model temporal priors and cross-sensor validation which provides a transferable blueprint for resilient perception in other safety-critical autonomous systems, such as unmanned aerial vehicle (UAV) collision avoidance.



\definecolor{BestBlue}{RGB}{0,92,184}      
\definecolor{GoodTeal}{RGB}{0,128,128}     
\definecolor{WarnOrange}{RGB}{217,95,2}    
\definecolor{GrayText}{gray}{0.35}

\newcommand{\appendixbest}[1]{\textcolor{BestBlue}{\textbf{#1}}}
\newcommand{\appendixgood}[1]{\textcolor{GoodTeal}{\textbf{#1}}}
\newcommand{\appendixwarn}[1]{\textcolor{WarnOrange}{\textbf{#1}}}
\newcommand{\appendixcapstrong}{\textcolor{BestBlue}{\checkmark\checkmark}}
\newcommand{\appendixcapmed}{\textcolor{GoodTeal}{\checkmark}}
\newcommand{\appendixcaplow}{\textcolor{GrayText}{--}}

\FloatBarrier

\end{document}